\pdfoutput=1

\documentclass[11pt]{article}

\usepackage[]{acl}

\usepackage{times}
\usepackage{latexsym}

\usepackage{graphicx}
\usepackage{multirow}
\usepackage{array}
\usepackage{booktabs}

\usepackage{bm}
\usepackage{color,colortbl}
\usepackage{stfloats}
\usepackage{subfigure}
\usepackage{enumitem}
\usepackage{booktabs}
\usepackage{nccmath}
\usepackage{multirow}
\usepackage{subfigure}

\usepackage{algorithm}
\usepackage{algpseudocode}

\usepackage[T1]{fontenc}

\usepackage[utf8]{inputenc}

\usepackage{microtype}

\title{Learning from Bootstrapping and Stepwise Reinforcement Reward:\\ A Semi-Supervised Framework for Text Style Transfer}

\author{Zhengyuan Liu, \ Nancy F. Chen \\
  Institute for Infocomm Research, A*STAR, Singapore \\
  \texttt{\{liu\_zhengyuan,nfychen\}@i2r.a-star.edu.sg}}
  
\date{}

\begin{document}
\maketitle

\begin{abstract}
Text style transfer is an important task in controllable language generation. Supervised approaches have pushed performance improvement on style-oriented rewriting such as formality conversion. However, challenges remain due to the scarcity of large-scale parallel data in many domains. While unsupervised approaches do not rely on annotated sentence pairs for each style, they are often plagued with instability issues such as mode collapse or quality degradation.
To take advantage of both supervised and unsupervised paradigms and tackle the challenges, in this work, we propose a semi-supervised framework for text style transfer. First, the learning process is bootstrapped with supervision guided by automatically constructed pseudo-parallel pairs using lexical and semantic-based methods. Then the model learns from unlabeled data via reinforcement rewards. Specifically, we propose to improve the sequence-to-sequence policy gradient via stepwise reward optimization, providing fine-grained learning signals and stabilizing the reinforced learning process. Experimental results show that the proposed approach achieves state-of-the-art performance on multiple datasets, and produces effective generation with as minimal as 10\% of training data.
\end{abstract}

\section{Introduction}
\label{sec:introduction}
Text style transfer is a task in natural language generation, which aims to automatically control certain attributes during sentence paraphrasing, such as formality, sentiment, and humor \cite{rao-tetreault-2018-GYAFC, li2018delete}. Style transfer has many practical applications, such as altering emotions of spoken utterances, removing biases in transcripts, and conveying politeness in messages \cite{hovy1987genStyle}.
The key for a successful rewrite is to preserve the semantic content of the source sentence, while transforming it to a particular target style without sacrificing fluency and grammatical accuracy. Therefore, the performance of style transfer models is commonly assessed on both style accuracy and content preservation. 
When large-scale annotated sentence pairs are available, training neural sequence-to-sequence models via supervised learning shows impressive generation quality \cite{rao-tetreault-2018-GYAFC,lai2021thankBart}. 
However, in many use cases, it is unfeasible to adopt supervised approaches because parallel samples are unavailable.
To address data insufficiency bottlenecks, various unsupervised approaches have been proposed for text style transfer, including learning disentangled representations of style and content \cite{Shen2017-CrossAlign} and adopting pairwise back-translation \cite{prabhumoye-2018-BackTranslate}. Recently, reinforcement learning (RL) is introduced to develop unsupervised models such that rewards of content preservation and style conversion are used to optimize sequence generation  \cite{fuli-2019-DualRL,gong-2019-RL}.
However, RL-based methods are often challenging to train in practice. For instance, the rewards have high variance during early stages when learning from scratch, which affects the training stability; and they cannot provide fine-grained learning signals as traditional token-level maximum likelihood estimation, since they are often calculated on the entire generated sequence \cite{d2019textGAN}. As a result, models are prone to mode collapse and often fail to produce acceptable generations in reality. 

Herein, we propose a semi-supervised framework for text style transfer, and optimize it on training stability and signal fineness. Our semi-supervised model uses a small amount of parallel data for supervised learning, and gets further improvement by learning from a large amount of unlabeled data.
In contrast to prior work that often relies on human-annotated parallel pairs like \cite{chawla-2020-semi}, the approach we propose bootstraps the training process with automatically constructed pseudo parallel data. Two pseudo pair matching methods are investigated: a lexical-based strategy, which is straightforward by calculating the token-level overlap; and a semantic-based strategy, which uses semantic similarity as criteria and would have better general potential.

Furthermore, to obtain fine-grained signals for the RL-based sequence-to-sequence training process, we propose a stepwise reward re-weighting strategy. This is inspired by the observation that the style transfer weights are not uniform across tokens/spans in the source sentence: some tokens weigh more during attribute-guided text style transfer \cite{li2018delete}.
Therefore, instead of using the reward (e.g., style strength scores) calculated from the entire generated sentence \cite{fuli-2019-DualRL,lai2021thankBart}, we use the token-level reward. Specifically, we extract attribute-related attentive scores from a pre-trained style discriminator, obtain a stepwise reward by re-weighting the sequence-level score, and utilize it as a fine-grained signal for policy gradient back-propagation.

We evaluate the proposed framework that incorporates both supervision and reward-based learning on three style transfer corpora (Section \ref{sec:experiments}). Experiments show that our model achieves state-of-the-art performance. Particularly, the proposed model can produce reasonable generations with only 10\% training data on the Yelp and Amazon corpora, and it also outperforms the supervised baselines when applying on the well-annotated GYAFC dataset.

\section{Related Work}
\label{sec:related_work}
\paragraph{Neural Text Style Transfer}
The aim of text style transfer is to automatically convert text to a certain style while preserving the content \cite{mcdonald1985computational,hovy1987genStyle}. It has many applications, like persona-based dialogue generation \cite{niu2018polite}.
Recently, neural sequence-to-sequence architectures becomes popular for this task. When parallel data are available, supervised training with cross-entropy loss is typically applied \cite{rao-tetreault-2018-GYAFC}. 
However, annotated data are hard to obtain in many use cases, thus learning from non-parallel corpora has become an active research area. There are two approaches: (1) Disentangling style and content by learning a distinct representation for each element. For example, variational autoencoders are first used to transform a sentence into a low-dimension hidden state. Then the attribute-related latent representation is extracted to guide the decoder for target style generation \citep{Shen2017-CrossAlign, Fu2018StyleTI, john-2019-Distangle}; (2) Back translation, which uses cyclic reconstruction to improve content preservation \citep{Zhirui-2018-UMT-Style, prabhumoye-2018-BackTranslate,lample2019-MultipleAttribute,fuli-2019-DualRL}. For model optimization, some studies focus on applying reinforcement learning (RL), which defines a reward from a style classifier or a reward from back-translation to enhance style strength and content preservation \cite{gong-2019-RL, fuli-2019-DualRL, wu-2019-HieraStyle, Abhilasha-2020-Rewards}.
Recently, large-scale pre-trained language models are introduced to improve generation quality \cite{radford-2019-GPT2}, and have been incorporated in both semi-supervised \cite{chawla-2020-semi} and supervised approaches \cite{lai2021thankBart}. In this work, we use the BART \cite{lewis2020bart} as our language model backbone.

\noindent\textbf{Pseudo Data Augmentation}
To tackle the data scarcity challenge in text style transfer, one solution is to build pseudo pairs from massive non-parallel data.
\citet{zhang-2020-parallelAug} proposed several augmentation methods for pre-training a Transformer-based model and fine-tuning on human annotations. \citet{wang-2019-harn} proposed using harness-rule-based pre-processing, and joint training of bi-directional transfer and auto-encoder with two auxiliary losses \cite{wang-2020-LatentStyle}. \citet{jin2019iterMatch} and \citet{nikolov2019largePairs} constructed the pseudo corpora by iteratively matching via cosine similarity of sentence embeddings and hierarchical alignment. In this work, we use pseudo data as weak-supervision to bootstrap the training process, and further combine it with RL-based learning.

\noindent\textbf{Attribute Salience Assessment} In template-based and prototype editing methods for text style transfer, attribute marker detection is used to label the salient words and spans \cite{li2018delete}. Aside from n-gram statistical features, neural attention-based methods train attribute-related classifiers, and consider words with attention weights higher than average as markers \cite{dzmitry-2015-attention,Jingjing-2018-UnpairedStyle,sudhakar2019transforming}. \citet{zhou2020exploring} use the attribute salient scores as one of the model prediction output. To the best of our knowledge, we are the first to employ token-level attribute salience scores for reward re-weighting on policy gradient for sequence generation, and prior work only focuses on using attribute markers for text manipulation such as token replacement and template construction \cite{niu2018polite}.

\section{Methodology}
\label{sec:methodology}

Define $S$ as the source style and $T$ as the target style (e.g., $S$ = $negative$, $T$ = $positive$). Let $\mathcal{D}_S$ and $\mathcal{D}_T$ be the two datasets which are comprised of sentences in each style respectively. 
The style transfer system, denoted as a text encoding-decoding model $G$, is to generate sentences in the target style. The goal is formulated to maximize $P(\bm{y}|\bm{x};\theta_G)$, where $\theta_G$ are the model parameters. 
In our setting, we make the rewriting bidirectional, i.e. it can be used to transfer source style to target style and verse versa. In this case, an additional input $\bm{c} \in \{S, T\}$ is fed to $G$ specifying the style to which the sentence is to be converted. Hence, the objective is to maximize $P(\bm{y}|\bm{x}, \bm{c}; \theta_G)$.

\begin{figure*}[ht!]
    \begin{center}
    \includegraphics[width=0.99\textwidth]{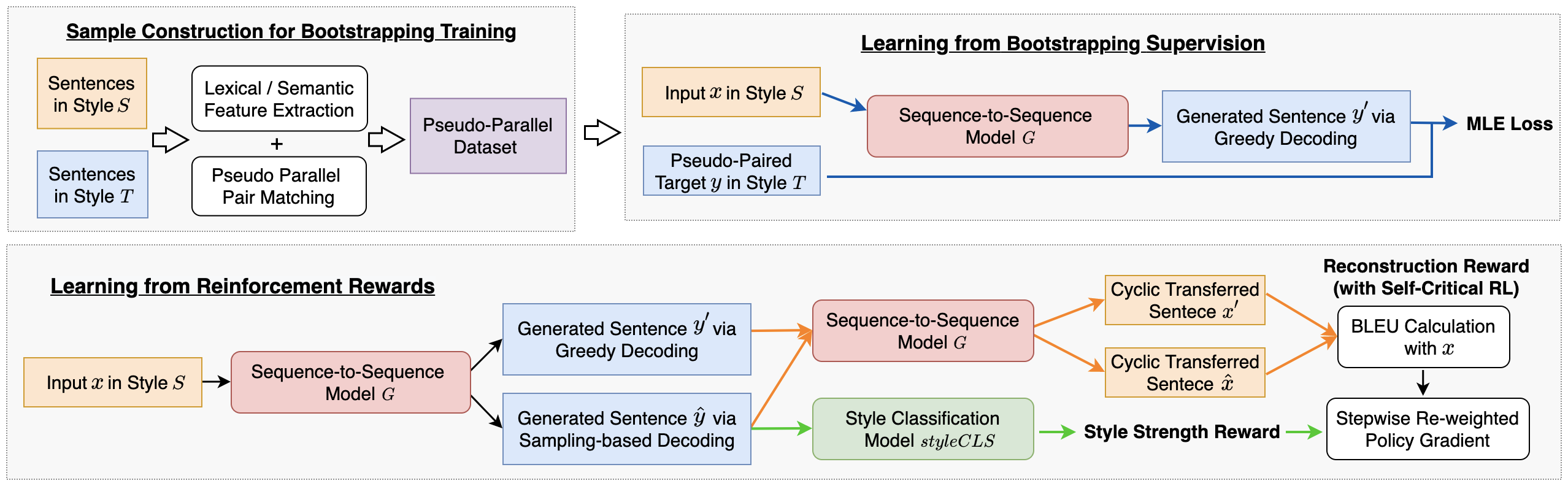}
    \end{center}
    \vspace{-0.2cm}
     \caption{Overview of the proposed framework. Text samples in two different styles are in yellow and in blue. The sequence-to-sequence model is shared by style transfer and cyclic generation. MLE loss, reconstruction reward, and style reward flows are in blue, yellow, and green arrow lines, respectively. See Algorithm \ref{alg:training} for training process.}
    \label{fig-framework}
\vspace{-0.2cm}
\end{figure*}

\subsection{Framework Overview}
\label{ssec:framework}
The overview of our proposed semi-supervised framework is shown in Figure \ref{fig-framework}. 
Given the non-parallel datasets $\mathcal{D}_S$ and $\mathcal{D}_T$, we use lexical or semantic features for pseudo parallel pair matching.
The training process consists of two stages: (1) the generator model $G$ is trained on the pseudo parallel samples, where cross-entropy loss over the target sentence tokens is used to optimize generated output probabilities, i.e. the \textbf{bootstrapping} step; (2) we
incorporate reconstruction and style rewards to enhance attribute rewriting and content preservation, where reinforcement learning is used to optimize the generation, i.e. the \textbf{reward-based learning}. Moreover, the second stage can use pseudo parallel pairs as well as the non-parallel samples. 

\subsection{Pseudo Parallel Data Construction}
\label{ssec:pseudo_construction}
To build the pseudo parallel data for bootstrapping, we investigate lexical similarity and semantic similarity for sentence matching.\\
\noindent\textbf{Lexical Similarity}
In text style transfer, rewriting is often accomplished by changing a few words or phrases that are indicative of a particular attribute in the source sentence, namely attribute markers, while leaving the rest largely unaltered \cite{li2018delete}. For example, \textit{``Moving past the shape, they were dry and truly tasteless.''}, a sentence with a negative sentiment style, can be transferred to a positive style by changing or replacing sentiment-specific words \textit{``dry''} and \textit{``tasteless''}, while keeping other words intact. This intuition has inspired the template-based and editing-based rewriting approaches \cite{li2018delete}. Here we employ it for the lexical feature extraction. First, from unaligned corpora of two styled subsets (e.g., positive, negative), we identify attribute markers by sorting phrases that occur 
with far higher frequency within one attribute than the other (e.g., \textit{``worst''} and \textit{``very disappointed''} are negative markers). Second, for each sentence in the two subsets, we remove those markers, and regard the remaining words as its content-preserved spans. Then we match the content-preserved spans of style $S$ to those of style $T$ with the smallest Levenshtein editing distance (see examples shown in Table \ref{table-pairs-example}).

\noindent\textbf{Semantic Similarity}
While the lexical features are straightforward and computationally-efficient, it may not generalize well in some tasks like formality conversion due to the ubiquitous span paraphrasing. Therefore, in this paper, we introduce semantic features for the pseudo data construction. While samples in different styles stand in different polarities, they are expected to be similar in the content-level semantic space. More specifically, for a sample $i$ in style $S$, we match it to the closest sentence in style $T$ in a semantic space. We use an unsupervised sentence representation model with contrastive learning \cite{gao2021-simCSE}, which achieves comparable performance to the supervised sentence embedding models, and calculate cosine similarity to measure the distance.\footnote{Additionally, we observed that in some corpora like Amazon \cite{li2018delete}, there are a number of samples labeled with incorrect style due to data noise, and the semantic approach is sensitive on this issue. Therefore, we use a style classifier to filter out the incorrectly clustered samples.} As shown in Table \ref{table-pairs-example}, the pseudo parallel data are similar at the semantic level, and they can be used as weak-supervision samples.

\begin{table}[t]
\linespread{1.0}
\centering
\small
\resizebox{1.0\linewidth}{!}
{
\begin{tabular}{p{7.8cm}}
\toprule
\textbf{Source Sentence:} if there were a way to put no stars, i would! \\
\textbf{Lexical Match:} i'd give it more stars if i could. \\
\textbf{Semantic Match:} love love love, if i could give you \_num\_ stars i would. \\
\midrule
\textbf{Source Sentence:} the manager sat us at our table, and she seemed very angry. \\
\textbf{Lexical Match:} the manager and employees are very nice. \\
\textbf{Semantic Match:} the manager alice herself came by our table and greeted us as well. \\
\midrule
\textbf{Source Sentence:} furthermore, i would rather drive \_num\_ minutes more to concord to race there. \\
\textbf{Lexical Match:} furthermore, they have a nice bar that goes both indoor and outdoor. \\
\textbf{Semantic Match:} i drive \_num\_ minutes to get here and it is definitely worth it! \\
\bottomrule
\end{tabular}
}
\caption{\label{table-pairs-example}Pseudo parallel sentence pairs extracted from Yelp sentiment transfer dataset. Source sentences are from the negative polarity set, and are matched to sentences from the positive set.}
\vspace{-0.3cm}
\end{table}

\subsection{Learning with Supervision}
\label{ssec:supervised_learning}
With the pseudo parallel data, we can conduct supervised learning with token-level maximum likelihood estimation (MLE). In our framework, we use a sequence-to-sequence neural network. Since the large-scale pre-trained language models boost the performance of various downstream tasks, we use BART \citep{lewis2020bart} as the language backbone, which is a denoising autoencoder with strong language generation capability. 
Given a source sentence $\bm{x}$ and a reference sentence $\bm{y}$, the cross-entropy loss is calculated between the decoder’s output and the reference sentence:
\vspace{-0.1cm}
\begin{equation}
\label{eq:mle-loss}
    L_{\mathrm{MLE}}=
    -\Sigma_{i}\textrm{log}(p(\bm{y}_{i}|\bm{y}_{1:i-1}, \bm{x},\bm{c};\theta_G))
\end{equation}
\vspace{-0.1cm}
Moreover, to avoid the generation becoming over-fitting to the pseudo parallel data, we add the label smoothing on the cross-entropy loss \cite{muller2019labelSmooth}, with the smoothing weight $\lambda$ = 0.15.

\subsection{Learning with Rewards}
\label{ssec:reward_learning}
Upon the supervised learning from the pseudo parallel data, the model can be further improved by unsupervised learning from the massive unlabeled data.
For the unsupervised stage, we adopt reinforcement learning, and use two rewards to enhance style rewriting and content preservation. 

\paragraph{Reconstruction Reward}
Back translation has proved effective to improve content preservation, we feed the transferred sentence to model $G$ for the backward rewriting, and calculate reconstruction reward on the cyclic generation.
Here we measure the reward based on BLEU \cite{papineni-2002-BLEU} score as in \cite{Abhilasha-2020-Rewards} to foster content preservation, and adopt policy gradient \cite{sutton1999policy} with Self-Critical Sequence Training to reduce the variance \cite{rennie2017selfCritical}:
\vspace{-0.1cm}
\begin{equation}
\label{eq-reward-bleu}
\begin{small}
    R_{cyclic}=
    \mathrm{score}(G(\bm{y}'),\bm{x})-\mathrm{score}(G(\bm{\hat{y}}), \bm{x})
    \end{small}
\end{equation}
where $\bm{x}$ is the backward target, $G(\bm{\hat{y}})$ is the back-translated output from greedy decoding generation $\bm{\hat{y}}$, and $G(\bm{y}')$ is the back-translated from sampling-based generation $\bm{y}'$ over a multi-nominal distribution. Noted that the $\mathrm{score}$ function can also be ROUGE and language model perplexity. The former is more suitable for summarization tasks; the latter needs additional computation.

\paragraph{Style Classification Reward}
\label{sub:cls-reward}
Aside from content preservation, we use a style strength reward to optimize the model.
We train a Transformer model for the binary style classification, and use it to evaluate how well the transferred sentence $\bm{y}'$ matches the target style. The style reward is $R_{style}$ defined as the classification score:
\begin{equation}
\label{eq-cls-softmax}
    p(s_{style}|\bm{y}') = \mathrm{softmax}(\mathrm{styleCLS}(\bm{y}', \phi))
\end{equation}
where $\mathrm{styleCLS}$ denotes the style classifier, $\phi$ are the parameters of the classifier, which are fixed during the training of the generation framework. $\bm{y}'$ is the generated sentence by sampling from the multi-nominal distribution at each step. Then, the reward-based learning is conducted via Policy Gradient \cite{sutton1999policy} back-propagation:
\begin{equation}
\label{eq:total-reward}
R=\lambda_{cyclic}R_{cyclic}+\lambda_{style}(R_{style} - \gamma)
\end{equation}
\vspace{-0.6cm}
\begin{equation}
\label{eq:gradient-policy}
    \nabla_{\theta_G}J=
    E[R\cdot\nabla_{\theta_G}log(P(\bm{y}'|\bm{x},\bm{c};\theta_G))]
\end{equation}
where $R$ is the sum of cyclic and style reward, $\bm{y}'$ is the generated sentence by sampling from the multi-nominal distribution at each step, $\theta_G$ are trainable parameters of the generator, the weight ratio $\lambda$ are added on cyclic and style reward separately, and $\gamma$ is a style reward penalty (see Table \ref{table:paramater_detail_appendix}).
The overall objectives for $\theta_G$ are the loss of the base model (Eq. \ref{eq:mle-loss}) and the policy gradient of RL rewards (Eq. \ref{eq:gradient-policy}).

\begin{figure}[t!]
    \begin{center}
    \includegraphics[width=0.48\textwidth]{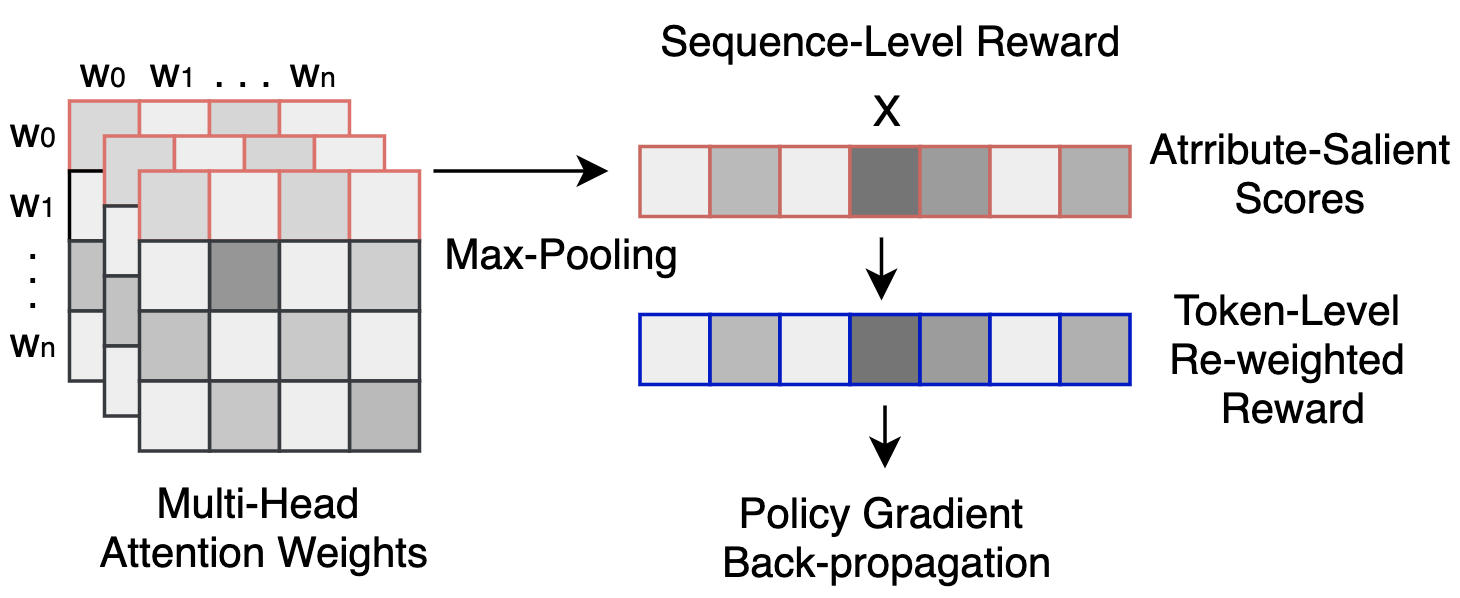}
    \end{center}
    \vspace{-0.3cm}
     \caption{Our proposed stepwise reward re-weighting.}
    \label{fig-stepwise}
\vspace{-0.4cm}
\end{figure}

\paragraph{Stepwise Reward Re-weighting}
\label{stepwise_reward} When applying reinforcement learning algorithms on sequence-to-sequence training, it is difficult for models to conduct end-to-end back-propagation due to the discrete nature of text. One of the common solutions is adopting policy gradient optimization \cite{sutton1999policy}, where the rewards are generally calculated on the whole output sequence. Since all generated tokens obtain the same reward value, this coarse-grained signal is suboptimal for learning performance and stability \cite{d2019textGAN}. For instance, when positive sentiment is targeted, the output sentence \textit{``I dislike this movie!''} will obtain a negative reward of style strength if its gold reference is \textit{``I love this movie!''}. In this context, the word \textit{``dislike''} should be punished more than the others in the sentence, but with sequence-level reward all words receive the same penalty. To address this drawback, we propose a solution by granulating the sequence-level reward with token-level salience scores, namely, stepwise re-weighting.

To re-weight the coarse-grained reward, we use the normalized attentive scores from the style classification model as the token-level attribute-salient scores. For the Transformer architecture, it is shown that heavily attended tokens correlate strongly with tokens that are indicative of the target style \cite{hewitt2019-Probe,vig2019analyzeBERT}.
Since the softmax linear layer is used over the attention stack of the first token $\langle$s$\rangle$ in a \textit{`RoBERTa-base'} model, the attention weights of other input tokens that correspond to $\langle$s$\rangle$ are of special interest in identifying significant sentence tokens. 
We inspect the attentions computed by the Transformer with 12 multi-head layers, and empirically observed that the attention weights of top layers correlate strongly with salient tokens (see the visualization in Appendix Figure \ref{fig-heatmap}). Given the attention matrix $A_i$ in the $i$-th multi-head layer, $a_{i}^j$ represents the attention vector of the first token (e.g., $\langle$s$\rangle$ , \textit{``[CLS]''}) from the $j$-th attention head, which is normalized across all tokens. We max-pool $A_i$ over all attention heads to form $a_i$, which represents the maximum extent to which each token was attended to by any head, and further max-pool the weights across the top-2 layers as the final stepwise attribute-salient scores (see layer selection in Section \ref{ssec:layer_select}), which are in the range of (0, 1). Then sequence-level rewards are expanded to the token length $n$, and re-weighted by the stepwise scores (see Figure \ref{fig-stepwise} and Algorithm \ref{alg:training}), and the policy gradient is formulated as following:
\vspace{-0.3cm}
\begin{equation}
\small
\label{eq:stepwise_gradient}
    \nabla_{\theta_G}J=
    E[\frac{1}{n}\sum_{t=1}^{n} R'_{t}\cdot\nabla_{\theta_G}log(P(\bm{y}'_{t}|\bm{y}'_{1:t-1},\bm{x},\bm{c};\theta_G))]
\end{equation}
\vspace{-0.5cm}

\begin{table}[t]
\centering
\small
\resizebox{\linewidth}{!}
{
\begin{tabular}{p{4.4cm}ccc}
\toprule
  Corpus & Train  & Valid  & Test \\
\midrule
  Yelp (Sentiment-Positive)   & 270K & 2,000 & 500 \\
  Yelp (Sentiment-Negative)   & 180K & 2,000 & 500 \\
\midrule
  Amazon (Sentiment-Positive)   & 277K & 985 & 500 \\
  Amazon (Sentiment-Negative)   & 278K & 1,015 & 500 \\
\midrule
  GYAFC E\&M (Formality-Paired)  & 52.6K & 2,877  & 1,416 \\
  GYAFC F\&R (Formality-Paired)  & 51.9K & 2,788 & 1,432 \\
\bottomrule
\end{tabular}
}
\caption{\label{table-data-stat}Statistics of the style transfer datasets. The GYAFC Entertainment\&Music (E\&M) and Family\&Relationships (F\&R) are comprised of paired samples. For Yelp and Amazon, only their test sets include human-written parallel references.}
\vspace{-0.4cm}
\end{table}

\begin{table*}[!ht]
\linespread{1.0}
\centering
\small
\resizebox{0.93\linewidth}{!}
{
\begin{tabular}{lp{1.1cm}<{\centering}p{1.1cm}<{\centering}p{1.1cm}<{\centering}p{1.1cm}<{\centering}p{1.3cm}<{\centering}}
\toprule
Model & Accuracy & BLEU & G2 & H2 & BertScore \\
\midrule
Cross Aligned \cite{Shen2017-CrossAlign}    &  75.3 & 17.9 & 36.7 & 28.9 & 68.3 \\
Back Translation \cite{prabhumoye-2018-BackTranslate}   &  95.4 & 5.0 & 21.9 & 9.6 & 61.0 \\
Style Embedding \cite{Fu2018StyleTI}     &  8.7 & 42.3 & 19.2 & 14.4 & 78.1 \\
Multi-Decoding \cite{Fu2018StyleTI}     &  50.2 & 27.9 & 37.4 & 35.9 & 69.4 \\
Unpaired \cite{Jingjing-2018-UnpairedStyle}   &  64.9 & 37.0 & 49.0 & 47.1 & 73.7 \\
Delete+Retrive \cite{li2018delete}  &  89.0 & 31.1 & 52.6 & 46.1 & 71.3 \\
Template-Based \cite{li2018delete}  &  81.8 & 45.5 & 61.0 & 58.5 & 73.7 \\
Unsupervised MT \cite{Zhirui-2018-UMT-Style}   &  95.4 & 44.5 & 65.1 & 60.7 & 80.8 \\
DualRL \cite{fuli-2019-DualRL}   &  85.6 & 55.2 & 68.7 & 67.1 & 84.1 \\
IterativeMatch \cite{jin2019iterMatch}   &  91.7 & 23.3 & 46.2 & 37.1 & 71.4 \\
Deep Latent \textbf{w/ Language Models} \cite{he2020-deepLatent}   &  85.2 & 46.4 & 62.8 & 60.0 & 76.4 \\
Direct Rewards \textbf{w/ GPT-2} \cite{liu2021directRewards}   &  91.2 & 53.8 & 70.0 & 67.6 & 83.6 \\
\midrule
Only Lexical Pseudo Data Bootstrapping (30K Pairs) &  81.3 & 26.5 & 46.4 & 39.9 & 72.1\\
Lexical Pseudo + Reward-Learning (30K)  &  81.1 & 50.4 & 63.9 & 62.1 & 82.1 \\
Lexical Pseudo + Reward-Learning (100K)   &  86.2 & 59.4 & \textbf{71.5} & \textbf{70.3} & \textbf{87.3} \\
\midrule
Only Semantic Pseudo Data Bootstrapping (30K Pairs) &  82.9 & 23.9 & 44.5 & 37.1 & 71.8 \\
Semantic Pseudo + Reward-Learning (30K)  &  83.5 & 49.6 & 64.3 & 62.2 & 82.5 \\
Semantic Pseudo + Reward-Learning (100K)   &  86.5 & 59.8 & \textbf{71.9} & \textbf{70.7} & \textbf{87.1} \\
\bottomrule
\end{tabular}}
\caption{\label{table-yelp-results}
Automatic evaluation scores on the Yelp sentiment style transfer task. Baseline results are reported with the model generations provided in published studies. Text examples are shown in Appendix Table \ref{table:generated_example_Yelp}.
}
\vspace{-0.2cm}
\end{table*}

\section{Experiments}
\label{sec:experiments}

\subsection{Experimental Datasets}
For extensive experiments, in this paper, we select three representative text style transfer corpora: Yelp (business reviews), Amazon (product reviews), and Grammarly’s Yahoo Answers Formality Corpus (GYAFC) \cite{li2018delete,rao-tetreault-2018-GYAFC}. The training, validation, and test split are the same as previous work \cite{fuli-2019-DualRL,chawla-2020-semi}, and their task types and statistics are shown in Table \ref{table-data-stat}. 
In the non-annotated corpora Yelp and Amazon, human-written references are only available for the test set. Therefore, to build the pseudo parallel data described in the previous section, we filter out the sentence pairs with lexical or semantic similarity lower than a threshold, and remove sentences that are shorter than 5 words. The pseudo parallel set is used for the bootstrapping training (Section \ref{ssec:supervised_learning}), and the rest samples are used for the unsupervised stage (Section \ref{ssec:reward_learning}).

\subsection{Experiment Setup}
The framework is implemented with Pytorch and Hugging Face Transformers\footnote{https://github.com/huggingface/transformers}. The \textit{`BART-base'} model is selected as the generator $G$. For style classification, \textit{`RoBERTa-base'} is used. We fine-tune models with AdamW \citep{diederik-kingma-2015-Adam} with batch size 32; initial learning rates are all set at $2e^{-5}$. Style reward penalty $\gamma$ is $0.2$. Values for $\lambda$ are set to $1.0$ for style reward and $0.8$ for cyclic reward. Beam search size is set at $6$.
Test results are reported with best validation scores (see Appendix Table \ref{table:paramater_detail_appendix} for environment and hyper-parameter setting details, and Algorithm \ref{alg:training} for the training process).

As previous work \cite{fuli-2019-DualRL,he-2020-UnsuperTrans,Abhilasha-2020-Rewards}, we adopt the following evaluation metrics: (1) \textbf{Style Accuracy} is calculated via binary classification to measure the style strength of re-writing. While the \textit{TextCNN} \citep{kim-2014-TextCNN} is used in previous studies, we also adopt a Transformer \textit{`RoBERTa-base'} classifier, where the reported scores are similar in our settings; (2) \textbf{BLEU} score is calculated on the prediction and human references to measure the content preservation; (3) We also compute the geometric mean (\textbf{G2}) and harmonic mean (\textbf{H2}) of style accuracy and BLEU score; (4) Since recent metrics with semantic similarity show better correlation with human judgments than traditional lexical measures. We also calculate \textbf{BertScore} between generation and references \cite{Zhang20BertScore}.

\subsection{Results on Yelp Corpus}
\label{ssec:result_yelp}
A number of representative unsupervised baseline models are selected for extensive comparison on the Yelp corpus: (1) models that adopt content-style disentanglement such as Cross Aligned \cite{Shen2017-CrossAlign} and Style Embedding \cite{Fu2018StyleTI}; (2) models that adopt back-translation such as Unsupervised MT \cite{Zhirui-2018-UMT-Style}, and Dual RL \cite{fuli-2019-DualRL}, and recent state-of-the-art models Deep Latent \cite{he2020-deepLatent} and Direct Rewards  w/ GPT-2 \cite{liu2021directRewards}. 
For our semi-supervised framework, we first (1) apply vanilla supervised learning to assess the effectiveness of the pseudo parallel data construction; (2) bootstrap the model with 30K pseudo parallel pairs, then further train it via reward-based learning; (3) apply semi-supervised learning by bootstrapping the model with 30K pseudo parallel pairs, and using 70K non-parallel samples for the reward-based training.
As shown in Table \ref{table-yelp-results}, vanilla supervised training on the 30K pseudo parallel data lead to favorable scores of style accuracy, though they do not perform well in terms of BLEU scores, as the pseudo pairs emphasize style converting rather than content preservation.
Further training with rewards improves both the style accuracy and BLEU score, and models with both lexical and semantic pseudo data produce comparable results with only 30k samples. Performance is further improved by using additional non-parallel data (70k samples), where our models outperform state-of-the-art baselines significantly.

\begin{table*}[!ht]
\linespread{1.0}
\centering
\small
\resizebox{0.90\linewidth}{!}
{
\begin{tabular}{lp{1.1cm}<{\centering}p{1.1cm}<{\centering}p{1.1cm}<{\centering}p{1.1cm}<{\centering}p{1.3cm}<{\centering}}
\toprule
Model & Accuracy & BLEU & G2 & H2 & BertScore \\
\midrule
Cross Aligned \cite{Shen2017-CrossAlign}	&	74.1	&	0.4	&	5.4	&	0.8 & 55.3	\\
Style Embedding \cite{Fu2018StyleTI} 	&	43.3	&	10.0	&	20.8	&	16.2 & 68.1	\\
Multi-Decoding \cite{Fu2018StyleTI} 	&	68.3	&	5.0	&	18.4	&	9.3 & 18.2	\\
Template-Based \cite{li2018delete} 	&	68.7	&	27.1	&	43.1	&	38.9	& 85.5 \\
Delete+Retrieve \cite{li2018delete} 	&	48.0	&	22.8	&	33.1	&	30.9 & 83.7 \\
Word-level Conditional GAN \cite{lai-etal-2019-wordGAN} 	&	77.4	&	6.7	&	22.7	&	12.3 & -	\\
Semi-LM-MMI \textbf{w/ BART-Large} \cite{chawla-2020-semi} 	&	68.9	&	28.6	&	44.4	&	40.4 & -	\\
Direct Rewards \textbf{w/ GPT-2} \cite{liu2021directRewards} 	&	68.3	&	38.6  &	51.3 & 49.3	&  72.1	\\
\midrule
Only Lexical Pseudo Data Bootstrapping (30K Data) &  79.8 & 16.4 & 36.1 & 27.2 & 63.3 \\
Lexical Pseudo + Reward-Learning (30K)  &  71.2 & 36.1 & 50.6 & 47.9 & 73.4 \\
Lexical Pseudo + Reward-Learning (100K)   &  73.1 & 46.3 & \textbf{58.1} & \textbf{56.6} & 78.4 \\
\midrule
Only Semantic Pseudo Data Bootstrapping (30K Data) &  81.2 & 10.3 & 28.9 & 18.2 & 60.5 \\
Semantic Pseudo + Reward-Learning (30K) &  72.3 & 35.5 & 50.6 & 47.6 & 72.7 \\
Semantic Pseudo + Reward-Learning (100K)   &  74.1 & 45.4 & \textbf{58.0} & \textbf{56.3} & 78.1 \\
\bottomrule
\end{tabular}}
\caption{\label{table-amazon-results}
Automatic evaluation scores on the Amazon sentiment style transfer task. Baseline results are calculated and reported with the model generations provided in published studies. See examples in Appendix Table \ref{table:generated_example_Amazon}.
}
\end{table*}

\begin{table*}[t!]
\linespread{1.0}
\centering
\small
\resizebox{0.98\linewidth}{!}
{
\begin{tabular}{lp{1.2cm}<{\centering}ccc|p{1.2cm}<{\centering}ccc}
\toprule
 &  \multicolumn{4}{c}{E\&M Domain} & \multicolumn{4}{c}{F\&R Domain} \\
Model & Accuracy* & BLEU & G2 & H2 & Accuracy* & BLEU & G2 & H2 \\
\midrule
Human Reference \cite{rao-tetreault-2018-GYAFC} & 81.5	&	100.0 & 90.2	&  89.8 &  80.5 & 100.0 & 89.7 & 89.2	\\
Rule-Based \cite{rao-tetreault-2018-GYAFC}	& 29.7	&	72.4	 &	46.4	& 42.1 & 82.1	&	65.8	 &	73.4	& 73.1	\\
Hybrid Annotations \cite{xu2019-hybrid}  & 28.8	&	69.2	 &	44.6 & 40.6 & 34.8	&	74.3	 &	50.8 & 47.3	\\
Semi-LM-MMI \textbf{w/ BART-Large} \cite{chawla-2020-semi}  & 30.4	&	76.5	 &	48.2 & 43.5 & 30.6	&	79.9	 &	49.4 & 44.2	\\
Rewarded \textbf{BART-Large} \cite{lai2021thankBart}  & 75.1	&	76.5	 &	75.7 & 75.7 & 74.6	&	79.2	 &	76.8 & 76.8	\\
\midrule
Only Labeled Data Supervision (Full)  &  75.0 & 71.2 & 73.1 & 73.1 &  73.7 & 72.5 & 73.1 & 73.1 \\
Labeled Data + Reward-Learning (30K)  &  75.7 & 71.4 & 73.5 & 73.4  &  72.4 & 74.4 & 73.3 & 73.4 \\
Labeled Data + Reward-Learning (Full)   &  82.2 & 71.0 & \textbf{76.3} & \textbf{76.2}  &  80.5 & 74.2 & \textbf{77.3} & \textbf{77.2} \\
\bottomrule
\end{tabular}}
\caption{\label{table-GYAFC-results}
Automatic evaluation scores on the GYAFC formality transfer task of baselines and our framework. Baseline results are reported with the generations provided as in \cite{chawla-2020-semi}. *The style accuracy is calculated with a fine-tuned \textit{`RoBERTa-base'} model (see Appendix for the result with \textit{TextCNN} classifier).
}
\vspace{-0.2cm}
\end{table*}

\subsection{Results on Amazon Corpus}
\label{ssec:result_amazon}
For the Amazon sentiment transfer corpus, we adopt the same training strategies described in Section \ref{ssec:result_yelp}. Aside from unsupervised models, we also select the semi-supervised model Semi-LM-MMI w/ BART \cite{chawla-2020-semi}, which adopted a language model-based discriminator for maximizing token-level conditional probabilities for training. Due to label noise in online-crawled data, the style accuracy for all models becomes lower than those trained on Yelp, and the classifier precision is only 86\% (see Table \ref{table-amazon-results}).
We also observed that the lexical similarity of pseudo parallel pairs is smaller than Yelp samples, and results in lower BLEU scores, especially when we apply supervised training on the 30K pseudo parallel data. On the other hand, content preservation largely benefits from the reward-based learning.
Unsurprisingly, after bootstrapping, training with rewards significantly improves the generation quality, and our framework achieves state-of-the-art performance. Moreover, bootstrapping with lexical-based and semantic-based pseudo data resulted in a similar final performance with reward learning.

\subsection{Results on GYAFC Corpus} 
\label{ssec:result_GYAFC}
In recent work, it is shown that style transfer models trained on parallel data can benefit from additional reward-based learning \cite{lai2021thankBart}. Here we conduct additional experiments to assess our semi-supervised framework on the GYAFC formality transfer corpus with well-annotated data.
We evaluate the proposed model on the \textit{informal}-to-\textit{formal} task as previous work \cite{chawla-2020-semi}, and compare them with strong baselines. 
As shown in Table \ref{table-GYAFC-results}, while the baselines show impressive BLEU scores on the formality transfer task, our framework outperforms them significantly in terms of style accuracy, approaching upper-bound human performance. Moreover, compared with the contemporary supervised work \cite{lai2021thankBart}, which also introduced additional RL-based optimization, our model still achieves higher G2 and H2 scores. The examples shown in Appendix Table \ref{table:generated_example_GYAFC} demonstrate that our approach  generates sentences with accurate formality paraphrasing.

\begin{table*}[!ht]
\linespread{0.95}
\centering
\small
\resizebox{0.88\linewidth}{!}
{
\begin{tabular}{lcccc|cccc}
\toprule
 &  \multicolumn{4}{c}{Yelp Data} & \multicolumn{4}{c}{Amazon Data} \\
Model & Accuracy & BLEU & G2 & H2 & Accuracy & BLEU & G2 & H2 \\
\midrule
Sequence-Level Reward (30K Data) &  85.1 & 26.5 & 47.5 & 40.4  &  78.4 & 19.0 & 38.5 & 30.5 \\
Stepwise Reward (30K Data)  &  81.1 & 50.4 & \textbf{63.9} & \textbf{62.1} &  71.2 & 36.1 & \textbf{50.6} & \textbf{47.9} \\
\midrule
Sequence-Level Reward (100K Data) &  84.8 & 35.3 & 54.7 & 49.8 & 81.4 & 21.9 & 42.2 & 34.5 \\
Stepwise Reward (100K Data) &  86.2 & 59.4 & \textbf{71.5} & \textbf{70.3} & 73.1 & 46.3 & \textbf{58.1} & \textbf{56.6} \\
\bottomrule
\end{tabular}}
\caption{\label{table-ablation-results}
Ablation study on the proposed stepwise reward on the Yelp and Amazon dataset. \textit{Sequence-level} denotes the reward is calculated on the whole sequence, without the stepwise re-weighting.
}
\vspace{-0.2cm}
\end{table*}

\subsection{Human Assessment}
\label{ssec:result_human_eval}

Additionally, we conducted a human evaluation on Yelp, Amazon and GYAFC datasets. Following previous work \cite{chawla-2020-semi,liu2021directRewards}, we evaluated the generated sentences from three aspects: style transfer strength (\textit{Style}), text fluency (\textit{Fluency}), and content preservation (\textit{Content}), separately. The three aspects are rated with range [1, 5], then their average value is calculated and reported as \textit{Mean} (see Table \ref{table-huma-eval-appendix} in Appendix). For each corpus, we randomly selected 80 test samples and compared the outputs of representative and previous state-of-the-art models. Each candidate was rated by three linguistic experts, and we report the average scores. Our model achieves better overall performance when considering all three evaluation metrics on each dataset. Moreover, we observe that leveraging the pre-trained language models such as BART and GPT-2 is beneficial for the text fluency.

\begin{table}[t!]
\linespread{0.95}
\centering
\small
\resizebox{0.9\linewidth}{!}
{
\begin{tabular}{p{1.4cm}p{1.1cm}<{\centering}p{0.9cm}<{\centering}p{0.7cm}<{\centering}p{0.7cm}<{\centering}}
\toprule
Layer No. & Accuracy & BLEU & G2 & H2 \\
\midrule
Layer-12 & 78.5 & 46.2 & 60.2 & 58.1 \\
Layer-11 & 81.1 & 45.5 & \textbf{60.7} & \textbf{58.2} \\
Layer-10 & 84.2 & 38.7 & 57.0 & 53.0 \\
Layer-9 & 72.3 & 43.8 & 56.2 & 54.5 \\
Layer-8 & 76.1 & 44.5 & 58.1 & 56.1 \\
Layer-7 & 70.3 & 41.6 & 54.0 & 52.2 \\
\bottomrule
\end{tabular}}
\caption{\label{table-layer-selection}
Layer selection for the proposed stepwise reward re-weighting. The Yelp sentiment transfer dataset and the semantic-based matching are used. We conduct experiments on the last 6 Transformer layers of the style classifier.
}
\vspace{-0.2cm}
\end{table}

\section{Analysis}
To extensively assess the effectiveness of the proposed methods, we conduct the following in-depth analyses.
\subsection{Ablation Study on Stepwise Re-weighting} 
We conduct an ablation experiment to assess the effectiveness of stepwise reward re-weighting. As shown in Table \ref{table-ablation-results}, the performance degrades significantly without the stepwise reward re-weighting, especially the BLEU score.
In particular, we observed that when removing stepwise optimization, the generator was prone to mode collapse. In one manifestation of mode collapse, the model appended a limited set of phrases to the source sentences, resulting in generation with disfluency and low diversity. It demonstrates that token-level reward optimization provides finer-granularity for policy gradient of sequence-to-sequence training. This approach can also be potentially extended to other text generation tasks.

\begin{table}[t!]
\linespread{0.95}
\centering
\small
{
\begin{tabular}{p{1.5cm}p{1.1cm}<{\centering}p{0.9cm}<{\centering}p{0.7cm}<{\centering}p{0.7cm}<{\centering}}
\toprule
Train Size & Accuracy & BLEU & G2 & H2 \\
\midrule
1,000 & 62.9 & 31.6 & 44.5 & 42.0 \\
5,000 & 68.2 & 36.8 & 50.0 & 47.8 \\
10,000 & 73.3 & 43.6 & 56.5 & 54.6 \\
15,000 & 76.1 & 45.5 & 58.8 & 56.9 \\
30,000 & 83.5 & 49.6 & 64.3 & 62.2 \\
\bottomrule
\end{tabular}}
\caption{\label{table-low-resource}
Results from different pseudo sample sizes using the proposed framework. The Yelp sentiment transfer dataset and semantic-based matching are used.
}
\vspace{-0.2cm}
\end{table}

\subsection{Attention Layer Selection for Stepwise Reward Re-weighting}
\label{ssec:layer_select}
We utilize attentive scores from the top-2 multi-head layers for stepwise reward re-weighting. To study the effect of layer selection, we compared the results using attention scores extracted from different Transformer layers in the style classifier described in Section \ref{ssec:reward_learning}. As shown in Table \ref{table-layer-selection}, the performance shows an overall increasing trend from the 7-th to the 12-th layer, and we obtained better results with the top layers. In scores of lower layers, we found that the model tended to assure content preservation rather than style accuracy. This is consistent with the observations from recent linguistic probing and model interpretation studies \cite{hewitt2019-Probe,xu2020understandBertSenti}: the information modeled in the Transformer-based networks, especially the pre-trained language backbones, is represented in a hierarchical manner, and the higher layers provide more effective information on scoring the span importance for text classification (see visualization in Appendix Figure \ref{fig-heatmap}).

\subsection{Bootstrapping Sample Size}
We investigate the effect of different pseudo parallel sample sizes. As shown in Table \ref{table-low-resource}, the result shows that the evaluation result by automatic metrics becomes acceptable when training reaches 10K samples. Results comparable to state-of-the-art are achieved with merely 30K data (10\% of the Yelp training set). We speculate that the relatively weak performance with 10K samples is because the BART model uses a denoising autoencoding paradigm \cite{lewis2020bart}, which is trained to reconstruct the input sentence, and style strength of sentence rewriting is strongly affected in this low resource scenario.

Additionally, we conduct an ablation study on the bootstrapping step, and the result shows that with the same training sample size, the generation performance (considering both style accuracy and content preservation) obtained significant improvement by adding the bootstrapping learning stage (see Appendix Table \ref{table-ablation-bootstrap}).

\section{Conclusions}
In this paper, we proposed a framework for text style transfer taking advantage of both supervised and unsupervised paradigms. The training process is bootstrapped with supervision guided by automatically constructed pseudo parallel data. Both lexical-based and semantic-based sentence matching proved effective. Moreover, the stepwise reward re-weighting significantly improved the generation performance, and is a generic design that can be easily extended. Experimental results showed that the proposed approach achieved state-of-the-art performance in multiple datasets, while producing reasonable generation even with minimal training data (10\% of original size).

\section*{Acknowledgments}
This research was supported by funding from the Institute for Infocomm Research (I2R) under A*STAR ARES, Singapore. We thank Ai Ti Aw for the insightful discussions. We also thank the anonymous reviewers for their precious feedback to help improve and extend this piece of work.

\bibliography{anthology,custom}
\bibliographystyle{acl_natbib}


\appendix

\section{Appendix}
\label{sec:appendix}

\begin{algorithm*}[h]
	\caption{Training process of the proposed semi-supervised text style transfer framework.}
	\label{alg:training} 
	\small
	\begin{algorithmic}[1]
    	\State {Given non-labeled datasets $\mathcal{D}_S$ and $\mathcal{D}_T$ in two different styles $S$ and $T$, construct pseudo parallel dataset $\mathcal{D}_{pseudo}$ with sentence pairs matched with lexical-based or semantic-based similarity}
    	\State {Pre-train a binary style classifier $\mathbf{styleCLS}$ on the two datasets $\mathcal{D}_S$ and $\mathcal{D}_T$}
		\State {Pre-train the text style transfer model $\boldsymbol{G_\theta}$ using pseudo-parallel sentence pairs in dataset $\mathcal{D}_{pseudo}$, with MLE loss (Eq.~\ref{eq:mle-loss}).}
    	\For{each iter $i=1,2,..., M$}
    		\State {Sample sentence $\boldsymbol{x}$ of source style $S$ from $\mathcal{D}_S$}
    		\State {Generate sentence $\boldsymbol{y}'$ of target style $T$ via model $\boldsymbol{G_\theta}$ by greedy decoding}
    		\State {Generate sentence $\boldsymbol{\hat{y}}$ of target style $T$ via model $\boldsymbol{G_\theta}$ by sampling-based decoding}
    		
    		\State \Comment{\textit{Reconstruction Reward Calculation (Content Preservation)}}
    		\State {Given $\boldsymbol{y}'$, generate back-translated sentence $\boldsymbol{x}'$ of source style $S$ via model $\boldsymbol{G_\theta}$ by greedy decoding}
    		\State {Given $\boldsymbol{\hat{y}}$, generate back-translated sentence $\boldsymbol{\hat{x}}$ of source style $S$ via model $\boldsymbol{G_\theta}$ by greedy decoding}
    		\State {Compute reconstruction reward $R_{cyclic}$ based on BLEU scores of the pair [$\boldsymbol{x}$, $\boldsymbol{x}'$] and the pair [$\boldsymbol{x}$, $\boldsymbol{\hat{x}}$], following Self-Critical Sequence Training (Eq.~\ref{eq-reward-bleu})}
    		
    		\State \Comment{\textit{Style Reward Calculation (Style Strength)}}
    		\State {Compute style reward $R_{style}$ of generated sentence $\boldsymbol{\hat{y}}$ using the style classifier $\mathbf{styleCLS}$}
    		
    		\State \Comment{\textit{Stepwise Reward Re-weighting}}
    		\State {Compute the stepwise re-weighting values by max-pooling attentive scores from style classifier $\mathbf{styleCLS}$ on the generated sentence $\boldsymbol{\hat{y}}$}
    		\State {Expand $R_{style}$ and $R_{cyclic}$ from 1-D (sequence level) to 2-D (token level), and re-weight $R_{style}$ with stepwise values}
    		\State {Compute the total stepwise reward $R'$ by adding $R_{style}$ and $R_{cyclic}$, based on Eq.~\ref{eq:total-reward}}
    		\State {Update $\boldsymbol{\theta}$ using reward $R'$ based on Eq.~\ref{eq:stepwise_gradient}}
    	\EndFor
	\end{algorithmic} 
\end{algorithm*}

\begin{table*}[h]
    \centering
    \small
    \resizebox{0.95\linewidth}{!}{
    \begin{tabular}{p{4.5cm}p{10cm}}
    \toprule
    \textbf{Environment Details} \\
    \midrule
        Sequence Generator & BART-Base (12-layer, 768-hidden, 16-heads, 139M parameters). \\
        Style Classifier & RoBERTa-base (12-layer, 768-hidden, 12-heads, 125M parameters). \\
        GPU Model & Single Tesla A100 with 40 GB memory; CUDA version 11.0. \\
        Library Version & Pytorch==1.8.1; Transformers==4.8.2. \\
        Computational Cost & Average 5 hours training time for one round. Average 3 rounds for each reported result (calculating mean of the result scores). \\
         \midrule
         \textbf{Hyper-parameter} &  \textbf{Setting Detail} \\
         \midrule
         Learning Rate and Batch Size & We set the learning rate and batch size according to regular language model fine-tuning strategy \cite{lewis2020bart}. \\
         Beam Search Size & We evaluated models on beam search sizes from 3 to 10, and 6 provided the best balance of performance and inference speed. \\
         Style Reward Penalty $\gamma$ (Eq. 4) & (1) In our experiment, we observed that the style reward $R_{style}$ values given by the style classifier were up to 0.9 (indicating a high level of style transfer strength), while the cyclic reconstruction reward $R_{cyclic}$ values were at a lower level (average was 0.5). Therefore, we added the $\gamma$ to adjust the $R_{style}$ to the same level of $R_{cyclic}$.
         (2) We evaluated values from 0.1 to 0.4 (0.1 as step), and empirically set the $\gamma$ at 0.2. Training without the penalty $\gamma$ did not produce significantly degraded results. \\
         $\lambda_{cyclic}$ and $\lambda_{style}$ (Eq. 4) & We evaluated both values with 1.0 +/- 0.2, and empirically set $\lambda_{cyclic}$ at 1.0,  $\lambda_{style}$ at 0.8. Setting at 1.0 by default did not produce degraded results.\\
         Sequence-Level \& Stepwise Reward & For the comparison of using sequence level and stepwise rewards, we run experiments with the aforementioned parameter setting. \\
         Combination of lexical and semantic pseudo-parallel data & In our pilot experiment, we tried to combine both lexical and semantic pseudo-parallel data, but this did not bring any improvement on the Yelp and Amazon. Presumably this is because the semi-supervised model only requires weak supervision from the pseudo-parallel data, and either the lexical and semantic data can provide sufficient information at the bootstrapping training stage.  \\
         \bottomrule
    \end{tabular}}
    \caption{The detailed environment settings and search strategy of training parameters in our experiment. It is worth mentioned that our proposed semi-supervised approach with bootstrapping strategy and stepwise reward re-weighting is targeted to tackle the unstable learning issue of RL-based models. }
    \label{table:paramater_detail_appendix}
\vspace{1.0cm}
\end{table*}

\begin{table*}[ht!]
\linespread{1.0}
\centering
\small
\resizebox{\linewidth}{!}
{
\begin{tabular}{lp{1.2cm}<{\centering}ccc|p{1.2cm}<{\centering}ccc}
\toprule
 &  \multicolumn{4}{c}{E\&M Domain} & \multicolumn{4}{c}{F\&R Domain} \\
Model & Accuracy* & BLEU & G2 & H2 & Accuracy* & BLEU & G2 & H2 \\
\midrule
Human Reference \cite{rao-tetreault-2018-GYAFC} & 58.7	&	100.0 & 76.6 & 73.9 & 51.4 & 100.0 & 71.6 & 67.8	\\
Rule-Based \cite{rao-tetreault-2018-GYAFC}	& 11.4	&	72.4	 &	28.7 & 19.6 & 52.1	&	65.8	 &	58.5 & 58.1	\\
Hybrid Annotations \cite{xu2019-hybrid}  & 10.4	&	69.2	 & 26.8 & 18.0 & 8.75	&	74.3	 &	25.4 & 15.6	\\
Semi-LM-MMI \textbf{w/ BART} \cite{chawla-2020-semi}  & 10.6	&	76.5	 &	28.4 & 18.6 & 9.68	&	79.9	 &	27.8 & 17.2	\\
Rewarded \textbf{BART-Large} \cite{lai2021thankBart}  & 52.8	&	76.5	 &	63.5 & 62.4 & 45.9	&	79.2	 &	60.2 & 58.1	\\
\midrule
Only Labeled Data Supervision (Full)  &  55.2 & 71.2 & 62.6 & 62.1 &  47.3 & 72.5 & 58.5 & 57.2 \\
Labeled Data + Reward-Learning (30K)  &  55.3 & 71.4 & 62.8 & 62.3  &  45.2 & 74.4 & 57.9 & 56.2 \\
Labeled Data + Reward-Learning (Full)   &  58.1 & 71.0 & \textbf{64.2} & \textbf{63.9}  &  50.3 & 74.2 & \textbf{61.0} & \textbf{59.9} \\
\bottomrule
\end{tabular}}
\caption{\label{table-GYAFC-TextCNN-appendix}
Automatic evaluation scores on the GYAFC formality style transfer task of baseline models and our framework. Baseline results are reported with the model generations provided in published studies \cite{chawla-2020-semi}. * The style accuracy is calculated with a \textit{TextCNN} classifier.
}
\end{table*}

\begin{table*}[ht!]
    \centering
    \small
    \resizebox{0.92\linewidth}{!}{
    \begin{tabular}{ll}
    \toprule
         \textbf{Model} &  \textbf{Text} \\
         \midrule
         Source Sentence & ever since joes has changed hands it 's just gotten worse and worse . \\
         Human Reference & ever since joes has changed hands it 's gotten better and better . \\
         Cross Aligned \cite{Shen2017-CrossAlign} & i recommend that has out to it 's always great and fun . \\
         Delete+Retrieve \cite{li2018delete}  & ever since joes has changed hands it 's just so good ! \\
         DualRL \cite{fuli-2019-DualRL} & ever since dedicated has changed hands it 's just gotten better and better . \\
         IterativeMatch \cite{jin2019iterMatch}  & dominos has gotten better and better . \\
         Deep Latent w/ LMs \cite{he2020-deepLatent} & just since their sausages has changed it 's just gotten worse and worse . \\
         Direct Rewards  w/ GPT-2 \cite{liu2021directRewards} & ever since joes has changed hands it 's just gotten better and better . \\
         Bootstrapping + Reward-Learning (Ours) & ever since joes has changed hands it 's just gotten better and better . \\
         \midrule
         Source Sentence & no , i 'm not at a scottsdale club . \\
         Human Reference & this was a great club. \\
         Cross Aligned \cite{Shen2017-CrossAlign} & great , i 'm so at a local business . \\
         Delete+Retrieve \cite{li2018delete}  & this is a great place to get a scottsdale club . \\
         DualRL \cite{fuli-2019-DualRL} & great job . \\
         IterativeMatch \cite{jin2019iterMatch}  & i 'm so glad i found this place . \\
         Deep Latent w/ LMs \cite{he2020-deepLatent} & great food , great service at a scottsdale club . \\
         Direct Rewards  w/ GPT-2 \cite{liu2021directRewards} & great , nice and a scottsdale club . \\
         Bootstrapping + Reward-Learning (Ours) & great , i 'm at a scottsdale club . \\
         \midrule
         Source Sentence & french toast plate was good , mom said , but eggs were cold . \\
         Human Reference & french toast plate was good , mom said , eggs were hot . \\
         Cross Aligned \cite{Shen2017-CrossAlign} & their food tasted was good , juicy , and fries are very clean . \\
         Delete+Retrieve \cite{li2018delete}  &  french toast plate was good , mom said , but eggs were amazing ! \\
         DualRL \cite{fuli-2019-DualRL} & french toast plate was good , mom said , but eggs were delicious . \\
         IterativeMatch \cite{jin2019iterMatch}  & the food was delicious and the eggs were fresh . \\
         Deep Latent w/ LMs \cite{he2020-deepLatent} & wow ! \\
         Direct Rewards  w/ GPT-2 \cite{liu2021directRewards} & french toast plate was good , mom said , with amazing eggs are warm . \\
         Bootstrapping + Reward-Learning (Ours) & french toast plate was good , mom said , but eggs were amazing . \\
         \midrule
         Source Sentence & however , it turned out to be nothing like i thought it would . \\
         Human Reference & this turned out exactly how i thought it would . \\
         Cross Aligned \cite{Shen2017-CrossAlign} & however , it right out to be great , it is the place . \\
         Delete+Retrieve \cite{li2018delete}  &  it turned out to be nothing like i thought it was so good ! \\
         DualRL \cite{fuli-2019-DualRL} & however , it turned out to be nothing extraordinary it would thought it would  \\
         IterativeMatch \cite{jin2019iterMatch}  & it turned out i worried about nothing . \\
         Deep Latent w/ LMs \cite{he2020-deepLatent} & loved it ! \\
         Direct Rewards w/ GPT-2 \cite{liu2021directRewards} & although , it turned out to be great with i thought it will . \\
         Bootstrapping + Reward-Learning (Ours) & however , it turned out to be great like i thought it would . \\
          \bottomrule
    \end{tabular}}
    \caption{Examples of human references and generated sentences on the Yelp corpus from representative baseline models and our proposed framework. The text style is converted from negative to positive.}
    \label{table:generated_example_Yelp}
\end{table*}

\begin{table*}[h]
    \centering
    \small
    \resizebox{0.94\linewidth}{!}{
    \begin{tabular}{ll}
    \toprule
         \textbf{Model} &  \textbf{Text} \\
         \midrule
         Source Sentence & it makes a buzzing sound when devices are plugged in. \\
         Human Reference & it makes a useful buzzing sound when devices are plugged in. \\
         Cross Aligned \cite{Shen2017-CrossAlign} & it s a nice , and easy to clean out . \\
         Style Embedding \cite{Fu2018StyleTI}  & it makes a bit different , while but num\_extend mode . \\
         Template-Based \cite{li2018delete}  & it makes a buzzing sound when devices are plugged in and use it to charge my . \\
         Delete+Retrieve \cite{li2018delete}  &  it makes a buzzing sound when the devices are plugged in . \\
         Direct Rewards  w/ GPT-2 \cite{liu2021directRewards} & it makes a cooking faster than devices are plugged in . \\
         Bootstrapping + Reward-Learning (Ours) & it makes a great sound when devices are plugged in . \\
         \midrule
         Source Sentence & it was not as good as our much cheaper model . \\
         Human Reference & its a great as before . \\
         Cross Aligned \cite{Shen2017-CrossAlign} & it s not not worth the phone and very well . \\
         Style Embedding \cite{Fu2018StyleTI}  & it was worth it size but at least my product , . \\
         Template-Based \cite{li2018delete}  & it was not as good as our much cheaper model and works just . \\
         Delete+Retrieve \cite{li2018delete}  &  as using the much cheaper model as it is also much cheaper . \\
         Direct Rewards  w/ GPT-2 \cite{liu2021directRewards} & it was excellent as our much cheaper model . \\
         Bootstrapping + Reward-Learning (Ours) & it was as good as our much cheaper model . \\
         \midrule
         Source Sentence & i received the wrong color and it shreds easily . \\
         Human Reference & i received the right color and it works well. \\
         Cross Aligned \cite{Shen2017-CrossAlign} & i bought the phone and it s easy to . \\
         Style Embedding \cite{Fu2018StyleTI}  & i received the fact that and quickly is no clean . \\
         Template-Based \cite{li2018delete}  & i received the wrong color and it shreds easily to order more . \\
         Delete+Retrieve \cite{li2018delete}  &  i received the wrong color and it looks very nice !  he would highly recommend it easily . \\
         Direct Rewards  w/ GPT-2 \cite{liu2021directRewards} & i received the best cooking efficiently . \\
         Bootstrapping + Reward-Learning (Ours) & i received the right color and it shreds easily . \\
         \midrule
         Source Sentence & i am actually afraid to open the remaining jars . \\
         Human Reference & I look forward to opening the remaining jars. \\
         Cross Aligned \cite{Shen2017-CrossAlign} & i have to say and the other ones . \\
         Style Embedding \cite{Fu2018StyleTI}  & i am actually used the right over a container .	 \\
         Template-Based \cite{li2018delete}  & i am actually afraid to open the remaining jars highly recommend . \\
         Delete+Retrieve \cite{li2018delete}  &  i am actually afraid to open the remaining jars this is great . \\
         Direct Rewards  w/ GPT-2 \cite{liu2021directRewards} & i am actually faster cooking than items . \\
         Bootstrapping + Reward-Learning (Ours) & i am actually happy to open the remaining jars . \\
          \bottomrule
    \end{tabular}}
    \caption{Examples of human references and generated sentences on the Amazon corpus from representative baseline models and our proposed framework. The text style is converted from negative to positive.}
    \label{table:generated_example_Amazon}
\end{table*}

\begin{table*}[h]
    \centering
    \small
    \resizebox{0.92\linewidth}{!}{
    \begin{tabular}{ll}
    \toprule
         \textbf{Model} &  \textbf{Text} \\
         \midrule
         Source Sentence & my dad likes action,my mom likes romance,but for me i like comedy. \\
         Human Reference & My father likes action, my mother likes romance, but for me I prefer comedy. \\
         Rule-Based \cite{rao-tetreault-2018-GYAFC}	& My dad likes action , my mom likes romance , but for me I like comedy . \\
         Hybrid Annotations \cite{xu2019-hybrid}	& My father likes action , my mother likes romance , but I like comedy . \\
         Semi-LM-MMI w/ BART-large \cite{chawla-2020-semi} & My dad likes action , my mom likes romance , but for me I like comedy . \\
         Rewarded BART-Large \cite{lai2021thankBart} & My dad likes action , my mom likes romance , but for me I like comedy . \\
         Labeled Data + Reward-Learning (Ours) & My father likes action, my mother likes romance, but for me I prefer comedy. \\
        \midrule
         Source Sentence & I want to be on TV! \\
         Human Reference & I would like to be on television. \\
         Rule-Based \cite{rao-tetreault-2018-GYAFC}	& I want to be on television ! \\
         Hybrid Annotations \cite{xu2019-hybrid} & I want to be on television . \\
         Semi-LM-MMI w/ BART-large \cite{chawla-2020-semi} & I want to be on TV . \\
         Rewarded BART-Large \cite{lai2021thankBart} & I would like to be on television. \\
         Labeled Data + Reward-Learning (Ours)	& I would like to be on television. \\
         \midrule
         Source Sentence & BUT IT IS OKAY TO KISS ON THE FIRST DATE. \\
         Human Reference & It is okay to kiss on the first date. \\
         Rule-Based \cite{rao-tetreault-2018-GYAFC}	& However, it is okay to kiss on the first date. \\
         Hybrid Annotations \cite{xu2019-hybrid} & It is okay to kiss on the first date . \\
         Semi-LM-MMI w/ BART-large \cite{chawla-2020-semi} & It is okay to kiss on the first date . \\
         Rewarded BART-Large \cite{lai2021thankBart}	& However, it is acceptable to kiss on the first date. \\
         Labeled Data + Reward-Learning (Ours)	& However, it is acceptable to kiss on the first date. \\
         \midrule
         Source Sentence & The same guy you wanna be in a relationship with? \\
         Human Reference & Do you want to be in a relationship with the same man? \\
         Rule-Based \cite{rao-tetreault-2018-GYAFC}	& The same man with whom you would like to be in a relationship? \\
         Hybrid Annotations \cite{xu2019-hybrid} & The same guy you want to be in a relationship with ? \\
         Semi-LM-MMI w/ BART-large \cite{chawla-2020-semi} & The same guy you want to be in a relationship with ? \\
         Rewarded BART-Large \cite{lai2021thankBart} & The same man you want to be in a relationship with ? \\
         Labeled Data + Reward-Learning (Ours)	& Is this the same man you want to be in a relationship with? \\
          \bottomrule
    \end{tabular}}
    \caption{Examples of human references and generated sentences on the GYAFC corpus from representative baseline models and our proposed framework. The text style is converted from informal to formal.}
    \label{table:generated_example_GYAFC}
\vspace{0.5cm}
\end{table*}

\begin{table*}[h]
\linespread{1.0}
\centering
\small
\begin{tabular}{p{6.5cm}p{1.3cm}<{\centering}p{1.3cm}<{\centering}p{1.3cm}<{\centering}p{1.3cm}<{\centering}}
I. Scoring result on the Yelp corpus\\
\\
\toprule
Model & Style & Fluency & Content & Mean \\
\midrule
Delete+Retrieve \cite{li2018delete} & 3.25 & 2.72 & 2.86 & 2.94 \\
IterativeMatch \cite{jin2019iterMatch} & 3.40 & 2.88 & 2.69 & 2.99 \\
Direct Rewards w/ GPT-2 \cite{liu2021directRewards} & 3.51 & 3.15 & 3.18 & 3.28 \\
Bootstrapping + Reward-Learning (Ours)  & 3.49 & 3.29 & 3.25 & \textbf{3.34} \\
\bottomrule
\\
\end{tabular}
\begin{tabular}{p{6.5cm}p{1.3cm}<{\centering}p{1.3cm}<{\centering}p{1.3cm}<{\centering}p{1.3cm}<{\centering}}
II. Scoring result on the Amazon corpus\\
\\
\toprule
Model & Style & Fluency & Content & Mean \\
\midrule
Template-Based \cite{li2018delete} & 2.78 & 2.36 & 2.55 & 2.56 \\
Delete+Retrieve \cite{li2018delete} & 2.94 & 3.08 & 2.73 & 2.91 \\
Direct Rewards w/ GPT-2 \cite{liu2021directRewards} & 3.20 & 3.23 & 2.21 & 2.88 \\
Bootstrapping + Reward-Learning (Ours)  & 3.31 & 3.28 & 3.12 & \textbf{3.23} \\
\bottomrule
\\
\end{tabular}

\begin{tabular}{p{6.5cm}p{1.3cm}<{\centering}p{1.3cm}<{\centering}p{1.3cm}<{\centering}p{1.3cm}<{\centering}}
III. Scoring result on the GYAFC corpus\\
\\
\toprule
Model & Style & Fluency & Content & Mean \\
\midrule
Hybrid Annotations \cite{xu2019-hybrid} & 2.56 & 3.15 & 3.13 & 2.95 \\
Semi-LM-MMI w/ BART \cite{chawla-2020-semi} & 3.12 & 3.47 & 3.22 & 3.27 \\
Rewarded BART-Large \cite{lai2021thankBart} & 3.36 & 3.60 & 3.33 & 3.43 \\
Labeled Data + Reward-Learning (Ours) & 3.37 & 3.67 & 3.37 & \textbf{3.47} \\
\bottomrule
\\
\end{tabular}

\caption{\label{table-huma-eval-appendix}
Human evaluation are conducted on the Yelp, Amazon, and GYAFC style transfer datasets. Following previous work \cite{chawla-2020-semi,liu2021directRewards}, we evaluated the generated sentences from three aspects: style transfer strength (\textbf{Style}), text fluency (\textbf{Fluency}), and content preservation (\textbf{Content}), separately. The three aspects are rated with range [1, 5], then their average value is calculated and reported as \textbf{Mean}. For each corpus, we randomly selected 80 test samples and compared the outputs of representative and previous state-of-the-art models. Each candidate was rated by three linguistic experts, and we report the average scores. Our model achieves better overall performance when considering all three evaluation metrics on each dataset. Moreover, we observe that leveraging the pre-trained language models such as BART and GPT-2 is beneficial for the text fluency.}
\vspace{-0.5cm}
\end{table*}

\begin{figure*}[h]
    \begin{center}
    \includegraphics[width=0.8\textwidth]{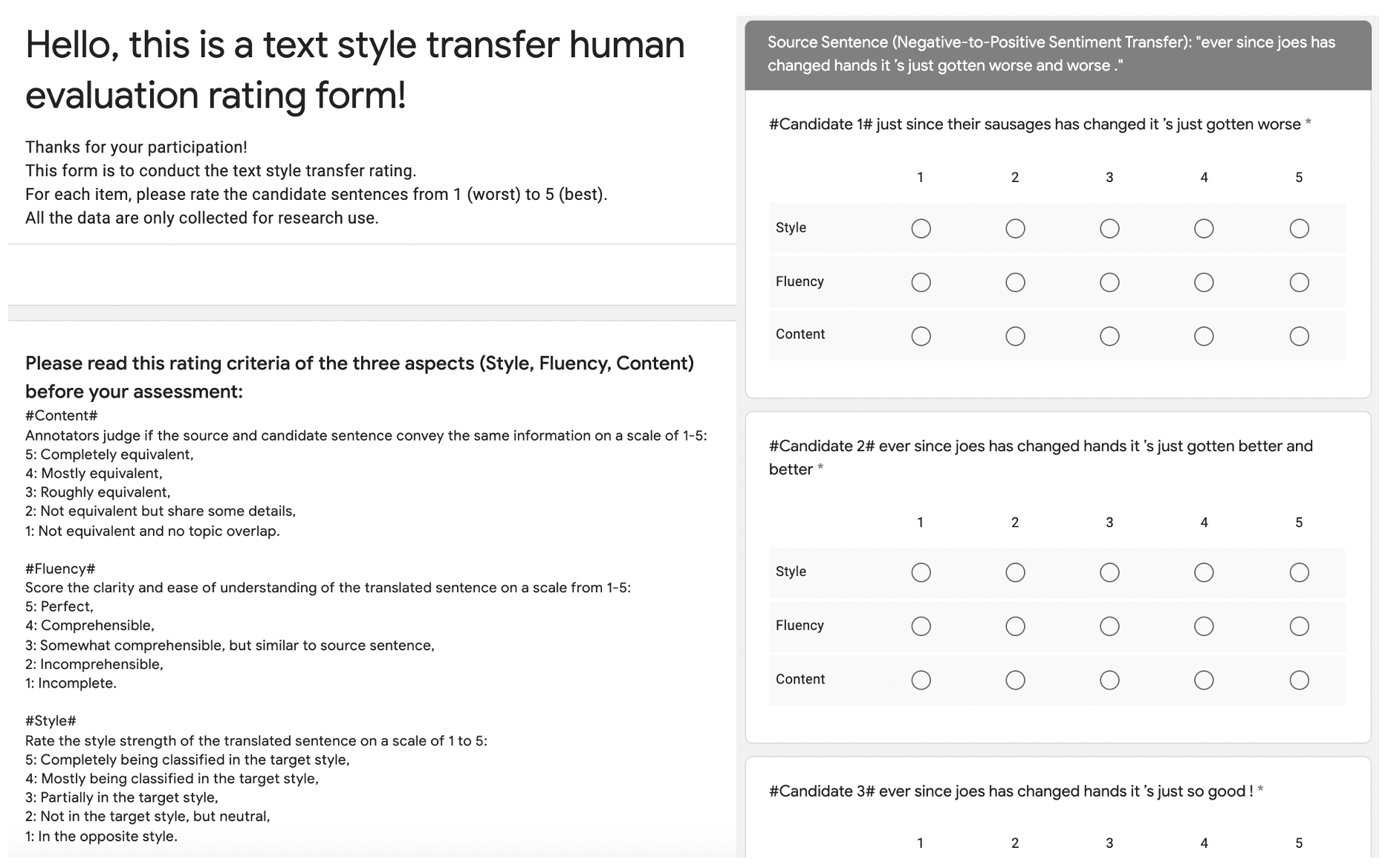}
    \end{center}
    \caption{Rating interface for the human evaluation. Text candidates are shuffled for each sample.}
    \label{fig-human-rating}
\end{figure*}

\begin{table*}[h]
\linespread{0.95}
\centering
\small
\resizebox{0.95\linewidth}{!}
{
\begin{tabular}{lcccc|cccc}
\toprule
 &  \multicolumn{4}{c}{Yelp Data} & \multicolumn{4}{c}{Amazon Data} \\
Model & Accuracy & BLEU & G2 & H2 & Accuracy & BLEU & G2 & H2 \\
\midrule
Lexical Pseudo + Reward-Learning (30K) &  81.1 & 50.4 & 63.9 & 62.1 & 71.2 & 36.1 & 50.6 & 47.9 \\
Pure Reward Learning (30K) &  70.8 & 41.3 & 54.0 & 52.1 & 61.2 & 26.1 & 39.9 & 36.5 \\
\midrule
Lexical Pseudo + Reward-Learning (100K)  &  86.2 & 59.4 & 71.5 & 70.2 & 73.1 & 46.3 & 58.1 & 56.6 \\
Pure Reward Learning (100K) &  75.5 & 46.1 & 58.9 & 57.2 & 65.6 & 26.5 & 41.6 & 37.7 \\
\bottomrule
\end{tabular}}
\caption{\label{table-ablation-bootstrap}
Ablation study of the proposed bootstrapping on the Yelp and Amazon datasets. Models are running in a RL-based unsupervised manner, and we used the same data sizes as the experiments in Table \ref{table-yelp-results} and Table \ref{table-amazon-results}.
}
\vspace{-1.0cm}
\end{table*}

\begin{figure*}[h!]
    \begin{center}
    \includegraphics[width=0.67\textwidth]{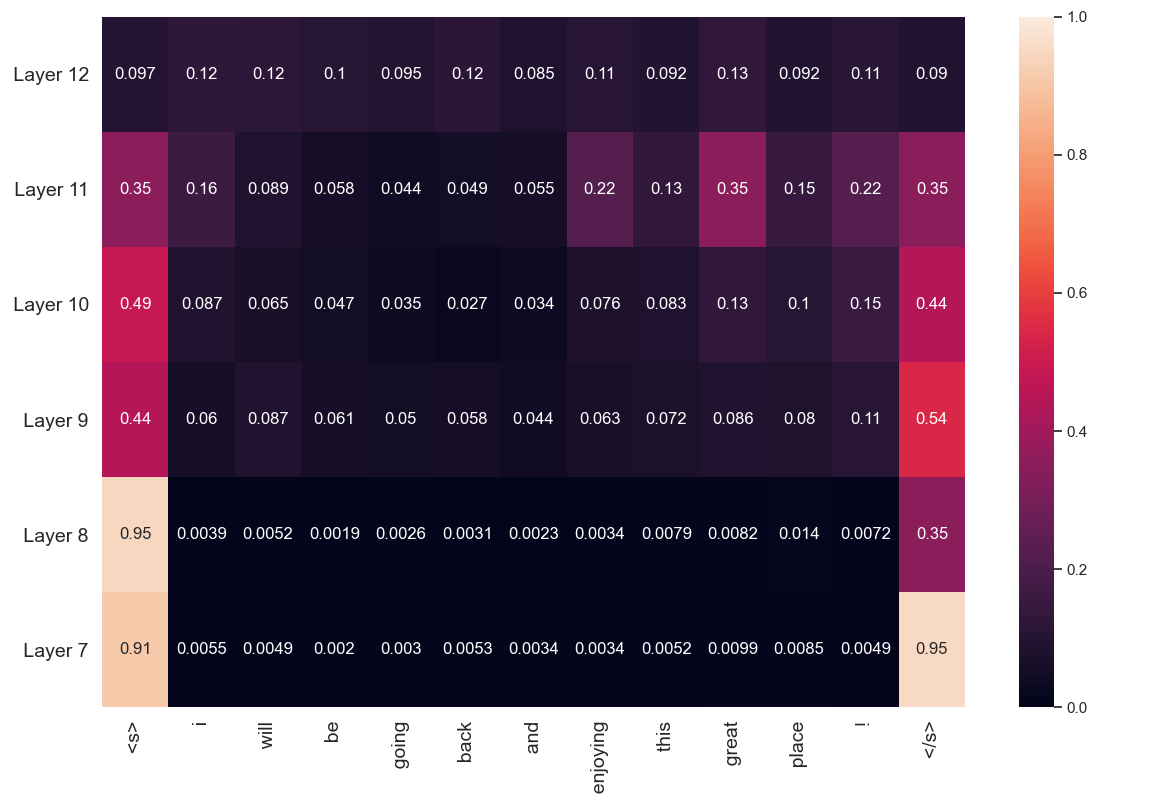}
    \includegraphics[width=0.67\textwidth]{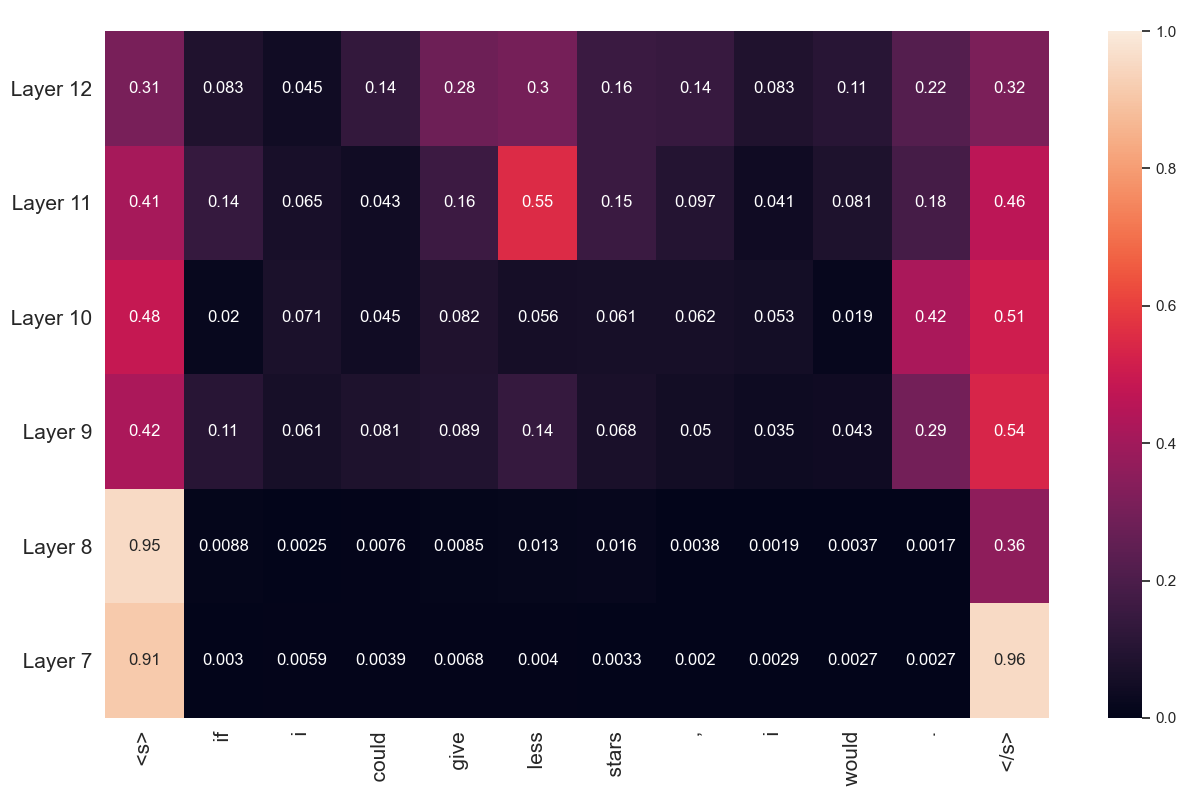}
    \end{center}
    \caption{Attention heatmap examples of the attention scores with layer-level max-pooling. The \textit{`RoBERTa-base'} model is fine-tuned on the Yelp data for style classification. The higher scores denotes higher attention weights on the tokens, and the top layers (especially the 11-th layer) shows better attribute-specified correlation. At the token level, the attention values and the max-pooled step-wise values described in Section \ref{ssec:reward_learning} are all in the range of (0, 1).}
    \label{fig-heatmap}
\vspace{-1.0cm}
\end{figure*}

\end{document}